%% file: main.tex
\documentclass[10pt,twocolumn,letterpaper]{article}
\PassOptionsToPackage{table,dvipsnames}{xcolor}
\usepackage[preprint]{cvpr}
\def\confName{CVPR}
\def\confYear{2027}

\usepackage{url}
\usepackage{todonotes}
\usepackage{amsmath, amsfonts, amssymb, amsthm}
\usepackage{graphicx}
\usepackage{caption, subcaption}
\usepackage{booktabs}
\usepackage{multirow}
\usepackage{wrapfig}
\definecolor{cvprblue}{rgb}{0.21,0.49,0.74}
\usepackage[pagebackref,breaklinks,colorlinks,allcolors=cvprblue]{hyperref}
\usepackage{amsmath}
\usepackage{amssymb}
\usepackage{verbatim}
\usepackage[ruled]{algorithm2e}
\usepackage{tikz}
\usetikzlibrary{calc, positioning, arrows.meta, shapes.geometric}
\usepackage[capitalize]{cleveref}
\Crefname{equation}{Eq.}{Eqs.}
\usepackage{multirow}
\usepackage[table]{xcolor}
\usepackage{enumitem}
\usepackage{adjustbox}
\usepackage{aliascnt}
\usepackage{makecell}
\usepackage{pifont}
\usepackage{balance}
\usepackage{todonotes}
\usepackage[dvipsnames]{xcolor}
\SetKw{KwOut}{Output:}{}
\SetKw{KwForward}{Forward pass:}{}
\SetKw{KwBackward}{Backward pass:}{}
\SetKw{KwInit}{Initialize:}{}
\SetKw{KwParameters}{Parameters:}{}
\newcommand{\w}{\mathbf{w}}
\newtheorem{proposition}{Proposition}
\input{files/table_style}

\title{CoDeQ: Compression with a Dead-Zone Quantizer for Differentiable Joint Pruning--Quantization}

\author{Jonathan Wenshøj\\
{\tt\small jowe@di.ku.dk}
\and
Mikkel Dahl\\
{\tt\small mikkel.dahl@di.ku.dk}
\and
Tong Chen\\
{\tt\small toch@di.ku.dk}
\and 
Bob Pepin\\
{\tt\small bob@pepin.io}
\and 
Raghavendra Selvan\\
{\tt\small raghav@di.ku.dk}\\
Department of Computer Science\\
University of Copenhagen
}

\begin{document}

\maketitle

\begin{abstract}
\input{files/abstract}
\end{abstract}
\input{files/introduction}
\input{files/related_work}
\input{files/background}
\input{files/method}
\input{files/experiments}
\input{files/discussion}

\input{files/conclusion}

% \clearpage
\section*{Acknowledgments}
% Authors acknowledge several people.
All authors acknowledge funding from the European Union’s Horizon Europe Research and Innovation Action program under grant agreements No. 101189771. RS acknowledges funding from the Independent Research Fund Denmark (DFF) under grant agreement No. 4307-00143B. The authors thank members of \href{https://saintslab.github.io/}{SAINTS Lab} for valuable discussions.
% \clearpage
\balance
{\small
\bibliographystyle{ieeenat_fullname}
\bibliography{references}
}
\onecolumn
\input{files/appendix}

% \clearpage
% NeurIPS-specific checklist retained in checklist.tex.
% \input{checklist}

\end{document}

%% file: files/table_style.tex
\newlength{\cquncwidth}
\providecommand{\cqresult}[2]{%
  \ensuremath{#1}%
  \makebox[\cquncwidth][l]{%
    \if\relax\detokenize{#2}\relax
    \else
      \ensuremath{\,{\scriptstyle\pm\,#2}}%
    \fi
  }%
}
\newsavebox{\cqtablebox}
\newcommand{\cqtabular}[3]{%
  \settowidth{\cquncwidth}{\ensuremath{\,{\scriptstyle\pm\,0.00}}}%
  \ifdefined\CodeQLayoutTest
    \sbox{\cqtablebox}{\begin{tabular}{@{}#2@{}}#3\end{tabular}}%
    \typeout{CODEQ-WIDTH #1: natural=\the\wd\cqtablebox; column=\the\columnwidth}%
    \ifdim\wd\cqtablebox>\columnwidth
      \PackageError{codeq-table-test}{Table #1 exceeds one column}{Shorten labels or adjust table spacing.}%
    \fi
  \fi
  \begin{tabular*}{\columnwidth}{@{\extracolsep{\fill}}#2@{}}#3\end{tabular*}%
}

%% file: files/abstract.tex
Joint pruning and quantization can reduce the inference cost of neural-networks, but determining layerwise sparsity and precision often requires auxiliary allocation procedures and considerable hyperparameter tuning. We introduce CoDeQ, a quantization-aware training method that unifies pruning and quantization through a learnable dead-zone quantizer. Its learned dead-zone width controls layerwise sparsity, while bit widths can be held fixed or learned in mixed-precision mode. A pruning-aware scale allocates all available nonzero reconstruction levels over the retained range. We derive the spacing that minimizes worst-case reconstruction error over this range for a fixed bit width and range endpoint. Model weights and compression parameters are optimized jointly under one global compression coefficient, with a shared learning rate for the compression parameters. 
On VGG7/CIFAR-10, CoDeQ with 2-bit activations achieves 0.05\% relative BOPs, for an accuracy drop of 0.32 percentage points. On DeiT-Tiny/ImageNet-1K, mixed-precision CoDeQ
preserves FP32-level accuracy at only 8.12\% relative BOPs.
Additional experiments across several architectures and datasets
demonstrate competitive accuracy--BOP trade-offs against
other joint compression methods.

%% file: files/introduction.tex
\section{Introduction}
\label{sec:introduction}
\begin{figure}[t]
    \centering
\includegraphics[width=0.65\columnwidth]{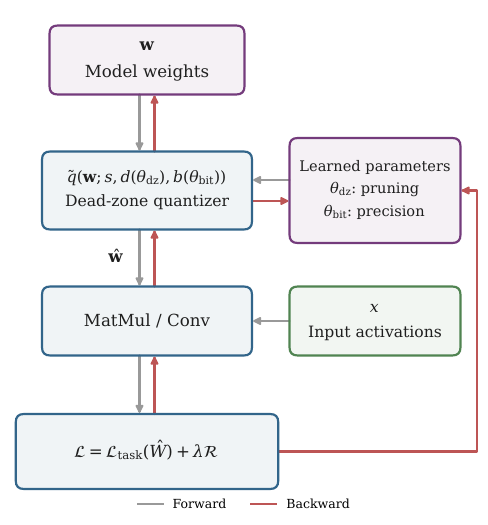}
    \caption{\textbf{CoDeQ Overview.}
    The learned dead zone induces weight sparsity, while nonzero
    reconstruction levels adapt to the retained range. Model weights
    and compression parameters are optimized using the task loss and
    compression penalty $\lambda\mathcal{R}$. Bit width is learned in
    mixed-precision mode. Gray and red arrows indicate forward
    computation and backward feedback, respectively.
    Layer indices are suppressed.}
    \label{fig:codeq_overview}
    \vspace{-0.5cm}
\end{figure}

Modern neural networks impose substantial memory and computational demands, creating barriers to deployment in resource-constrained environments~\cite{bartoldson2023compute}. Pruning and quantization address these demands by removing weights or structures and reducing the precision of the remaining values. Applying them sequentially separates decisions about which weights to retain from decisions about how to represent them, often requiring separate fine-tuning stages. Joint pruning--quantization brings these decisions into a shared optimization process, allowing sparsity and precision to be optimized together.

Although joint optimization opens the possibility of a single compression stage, many methods rely on an auxiliary procedure to determine layerwise sparsity and precision. Implementing this procedure, configuring its parameters, and adapting it to different architectures introduce practical challenges.

% \paragraph{Dependence on auxiliary procedures.}

Many joint methods {\bf depend on auxiliary procedures} to determine layerwise allocations through search, constrained projections, or heuristic updates~\cite{joint:CLIP-Q,joint:SQL,joint:QST,joint:qu2025automatic,joint:FITCompress,joint:APQ}. The user must implement this additional logic and integrate it with training, whether it runs as a separate stage or alternates with gradient updates. Differentiable alternatives such as DJPQ and Bayesian Bits learn compression parameters alongside network weights, avoiding auxiliary allocation procedures~\cite{joint:wang2020djpq,joint:van2020bayesian}. However, their formulations introduce additional gating mechanisms and optimization choices, while reported sparsity remains modest in some settings.

Auxiliary procedures also introduce settings beyond those specifying the compression target, including search budgets, candidate precisions, and allocation schedules ~\cite{joint:CLIP-Q,joint:SQL,joint:QST,joint:qu2025automatic,joint:FITCompress,joint:APQ}. Published recipes further vary compression weights and parameter-specific learning rates across models~\cite{joint:CLIP-Q,joint:wang2020djpq,joint:FITCompress}. Choosing these settings can require repeated experiments for each model--dataset pair, increasing the effort needed to obtain an effective accuracy--compression trade-off. 
Compression methods may also encode assumptions about supported layer types, model architecture, or relationships between layerwise compression decisions~\cite{joint:APQ,joint:QST,joint:wang2020djpq,joint:van2020bayesian,joint:FITCompress,joint:CLIP-Q}. These assumptions can limit how directly a procedure transfers to a new architecture. Applying the method to a different model may therefore require changes to the allocation logic itself.

To address these challenges, we introduce \textbf{Co}mpression with a \textbf{De}ad-Zone \textbf{Q}uantizer (\textbf{CoDeQ}). CoDeQ learns layerwise sparsity by parameterizing the quantizer's zero-bin width, without auxiliary allocation procedures. A single global compression coefficient and a shared compression learning rate govern its compression parameters. The weightwise operator preserves tensor dimensions, allowing the same formulation to apply to convolutional and linear layers without structural dependency analysis. The algorithm is illustrated in \Cref{fig:codeq_overview}. Our main contributions are:
\begin{itemize}[leftmargin=*]
    \item \textbf{A learnable quantizer for joint compression.}
    We parameterize the zero-bin width and bit width of a dead-zone quantizer and develop a QAT algorithm that learns these controls jointly with the weights under one global compression coefficient. We derive a pruning-aware scale that allocates the nonzero quantization levels over the retained range. This supports fixed-bit and mixed-precision training without auxiliary allocation procedures (\Cref{sec:forward,sec:objective,sec:backward}).

    \item \textbf{Packed inference with fused decoding.}
    We adapt existing dense and unstructured-sparse graphics processing unit (GPU) kernels~\cite{frantar2024marlin,xia2024flashllm,fan2025spinfer} to fuse decoding of packed CoDeQ weights with matrix multiplication using 16 bit floating-point (FP16) computation (\Cref{sec:kernel_experiments}).

    \item \textbf{Empirical evaluation and controlled studies.}
    We evaluate accuracy--bit operation (BOP) trade-offs on convolutional and transformer models, including quantized activations. Controlled studies examine compression strength, learned allocations, and nonzero-level allocation (\Cref{sec:experiments}).
\end{itemize}

%% file: files/related_work.tex
\section{Related work}
\label{sec:related_work}

\textbf{Joint pruning--quantization.}
Deep Compression~\cite{DBLP:journals/corr/HanMD15} applies pruning and quantization sequentially. Joint methods account for their interaction when selecting sparsity and precision. CLIP-Q~\cite{joint:CLIP-Q} selects layerwise policies through Bayesian optimization, while APQ~\cite{joint:APQ} uses a supernet and accuracy predictor to search architecture, pruning, and precision. FITCompress~\cite{joint:FITCompress} plans compression steps using Fisher-based sensitivity estimates. SQL~\cite{joint:SQL} alternates weight updates with budget-constrained projections, and QST~\cite{joint:QST} derives allocations heuristically from global targets. GETA~\cite{joint:qu2025automatic} combines projected optimization with a quantization-aware dependency graph, supporting CNNs and transformers. CoDeQ instead learns compression parameters through task and compression gradients within QAT, avoiding auxiliary allocation procedures altogether.\\
\textbf{Differentiable joint compression.}
DJPQ learns channel sparsity and quantizer parameters, while Bayesian Bits (BB) learns precision and pruning through a residual quantization hierarchy~\cite{joint:wang2020djpq,joint:van2020bayesian}. Both learn these compression parameters alongside network weights, but retain method-specific tuning and architectural handling. DJPQ's published recipes vary two regularization strengths and separate pruning/quantization learning-rate multipliers across models. BB specifies initialization and optimizer settings for its compression parameters, with additional fine-tuning on ImageNet. Their evaluated pruning constructions also depend on channel organization; BB additionally modifies its cost accounting and regularization for residual connections. Applying these methods therefore requires configuring their compression-specific training recipes and accommodating architectural dependencies.

BB also reports forward {\em storage equivalent to five model copies} in its standard precision hierarchy; the authors use checkpointing to alleviate the resulting memory pressure. Both methods report settings with pruning concentrated in relatively few layers while others remain largely dense~\cite{joint:wang2020djpq,joint:van2020bayesian}.

CoDeQ targets simpler adoption: one global compression coefficient and a shared compression learning rate govern a weightwise quantizer applicable to convolutional and linear layers. Preserving tensor dimensions avoids structural dependency analysis and compression-specific model restructuring.\\
\textbf{Parameterized quantizers.}
LSQ learns step sizes, while PACT learns activation clipping thresholds~\cite{esser2020learned,choi2018pact}. CoDeQ follows this tradition of parameterizing the quantization rule by learning the zero-bin width. This controls sparsity while allocating all available nonzero reconstruction levels over the retained range.\\
\textbf{Packed execution.}
Marlin, Flash-LLM, and SpInfer provide dense and unstructured-sparse GPU matrix multiplication~\cite{frantar2024marlin,xia2024flashllm,fan2025spinfer}. Our adaptations fuse decoding of packed CoDeQ codes with FP16 computation.

%% file: files/background.tex
\section{Background}
\label{sec:background}
\paragraph{Quantization.}
For weights $\w\in\mathbb R^n$ and bit width $b\in\{2,3,\ldots\}$, an elementwise symmetric uniform quantizer uses integer codes $\bar\w$ and quantized weights $\hat\w$:
\begin{equation}
\begin{aligned}
 \bar\w&=\operatorname{clip}(\left\lfloor\w/s\right\rceil,-Q_b,Q_b),\\
 \hat\w&=s\bar\w=q(\w;s,b),
\end{aligned}
\label{eq:quantizer_uniform}
\end{equation}
with $Q_b=2^{b-1}-1$. Here $s>0$ is the step size, $\lfloor\cdot\rceil$ rounds to the nearest integer with ties to even, and clipping limits the codes to $[-Q_b,Q_b]$. Weights with $|w_i|<s/2$ map to zero.
\paragraph{Magnitude pruning.}
Unstructured magnitude pruning applies the mask $m_i(\tau)=\mathbf1_{\{|w_i|>\tau\}}$ through $\mathbf m(\tau)\odot\w$, where $\odot$ denotes elementwise multiplication. It retains weights whose magnitudes exceed $\tau$ and sets the rest to zero.

\paragraph{Quantization-aware training.}
Quantization-aware training (QAT) retains full-precision latent weights for optimization while evaluating the task loss $\mathcal L_{\rm task}(q(\w;s,b))$ with quantized weights~\cite{jacob2017quantization,krishnamoorthi2018quantizing}.

\paragraph{Straight-through estimator.}
Rounding operator has zero derivative almost everywhere, blocking gradient flow in \Cref{eq:quantizer_uniform}. A straight-through estimator (STE)~\cite{bengio2013ste} replaces this derivative during backpropagation while preserving the hard forward operation. For \Cref{eq:quantizer_uniform}, an identity STE for rounding with the standard clipping derivative gives, at fixed $s$,
\begin{equation}
 \widetilde\partial_{w_i}\mathcal L_{\rm task}
 =\frac{\partial\mathcal L_{\rm task}}{\partial\hat w_i}
   \mathbf1_{\{|w_i|\le sQ_b\}}.
\label{eq:ste-quant-grad}
\end{equation}
We use $\widetilde\partial$ to distinguish a surrogate backward gradient from an ordinary derivative.

\paragraph{Bit operations (BOPs).}
To evaluate pruning and quantization jointly, we combine weight sparsity and precision in an idealized arithmetic-cost metric following~\cite{joint:wang2020djpq}. Let $\mathrm{MACs}_\ell$ be the dense multiply--accumulate (MAC) count of layer $\ell$, $p_\ell$ its measured nonzero fraction of quantized weights, and $b_\ell,a_\ell$ its weight and activation bit widths, respectively. The weight sparsity then is $1-p_\ell$. For unstructured weight sparsity we count
\begin{equation}
 \mathrm{BOPs}_\ell=\mathrm{MACs}_\ell \times p_\ell \times b_\ell \times a_\ell,
\label{eq:bops_definition}
\end{equation}
with $a_\ell=32$ for full-precision activations. Relative BOPs compare the counted layers with the same dense FP32 baseline:
%\begin{equation}
$
 \mathrm{Rel.\ BOPs}(\%)
 =100\,\frac{\sum_\ell \mathrm{BOPs}_\ell}
                  {32^2\sum_\ell \mathrm{MACs}_\ell}.
                  $
%\label{eq:relative_bops}
%\end{equation}

%% file: files/method.tex
% Background/method editorial revision: 12 November 2026.
% Keep notation helpers and legacy equation aliases local to this section.
\begingroup
\providecommand{\clip}{\operatorname{clip}}
\providecommand{\sign}{\operatorname{sign}}
\providecommand{\relu}{\operatorname{ReLU}}
\providecommand{\sg}{\operatorname{stopgrad}}
\providecommand{\norm}[1]{\left\lVert#1\right\rVert}

\makeatletter
\let\codeq@base@label\cref@label
\def\cref@label#1{%
  \codeq@base@label{#1}%
  \@ifundefined{codeq@aliases@#1}{}{\csname codeq@aliases@#1\endcsname}%
}
\newcommand{\codeqaliases}[2]{%
  \expandafter\def\csname codeq@aliases@#1\endcsname{%
    \@for\codeq@alias:=#2\do{%
      \expandafter\codeq@base@label\expandafter{\codeq@alias}}}%
}
\codeqaliases{eq:operator}{eq:dz_uniform_adjustable}
\codeqaliases{eq:assignment}{eq:variational_codeq}
\codeqaliases{eq:minimax}{eq:minimax_retained}
\codeqaliases{eq:range_coordinate}{eq:retained_range_coordinate}
\codeqaliases{eq:bit_coordinate}{eq:learnable_bit}
\codeqaliases{eq:adjacent}{eq:adjacent_bit_difference}
\codeqaliases{eq:objective}{eq:codeq_loss,eq:qpat_loss}
\codeqaliases{eq:regularizer}{eq:retained_span_cost}
\codeqaliases{eq:combined_gradients}{eq:weight_task_gradient}
\makeatother

\section{Method}
\label{sec:method}
CoDeQ learns compression within QAT by making the weight quantizer's zero-bin width adjustable. Its construction starts from the observation that the zero bin of a scalar quantizer implements magnitude pruning. A symmetric quantizer $\tilde q(\w;s,d,b)$ with zero-bin width $d$ obeys $\tilde q(\w;s,d,b)=\tilde q(\w;s,d,b)\odot\mathbf m(d/2)$.
The zero bin determines which weights are pruned, while nonzero reconstruction levels represent the retained weights. In a uniform quantizer we have $d=s$, so stronger pruning also increases the step size between nonzero levels. CoDeQ makes the zero-bin width adjustable and allocates the available nonzero levels over the retained range.

We first construct the quantizer and establish its properties, then describe how to learn its compression parameters under a joint compression objective using surrogate gradients.

\subsection{Joint pruning--quantization operator}
\label{sec:forward}
\label{sec:deadzone_quantizer}
Let $\w\in\mathbb R^n$ be a layer's latent weights, and let $W$ and $\hat W$ denote the collections of latent and quantized weights across the network. We suppress the layer index outside sums over layers and the algorithm. For bit width $b\ge2$, let $d/2\ge0$ be the pruning threshold and $s>0$ the step size. Uniform rounding itself introduces a zero bin. To prevent this from extending the intended pruning threshold beyond $d/2$, we compensate for its half-step width by setting \mbox{$\delta=d/2-s/2$}. We quantize and reconstruct weights as
\begin{equation}
\begin{aligned}
 \bar\w&=\clip\!\!\left(
 \left\lfloor\frac{\sign(\w)\!\odot\!\relu(|\w|\!-\!\delta)}{s}\right\rceil,
 -Q_b,Q_b\right),\\
 \hat\w&=\sign(\bar\w)\,\delta+s\,\bar\w
       =\tilde q(\w;s,d,b).
\end{aligned}
\label{eq:operator}
\end{equation}
All operations are elementwise with $\sign(0)=0$. The encoder shifts the magnitudes before rounding, so the zero/nonzero boundary lies at $\delta+s/2=d/2$. Consequently, $\hat w_i=0$ exactly when $|w_i|\le d/2$. The decoder reverses the shift for nonzero codes; multiplying the offset by $\sign(\bar\w)$ preserves zero codes exactly.
Thus $d$ controls the zero-bin width while $s$ sets the step size between nonzero levels. Setting $d=s$ recovers uniform quantization.
Let $R>0$ denote the chosen range endpoint. \Cref{fig:quantizer_comparison} contrasts uniform quantization, magnitude pruning followed by uniform quantization (P+U), and CoDeQ at the same $R$ and $b$.
\par\medskip\noindent
\begin{minipage}{\columnwidth}
\centering
\includegraphics[width=\columnwidth]{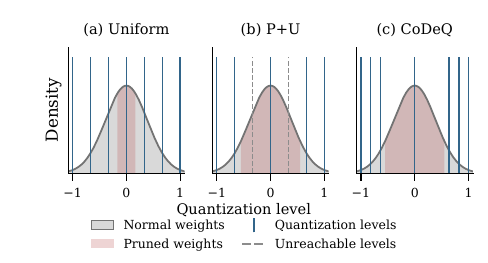}
\captionof{figure}{Quantizers with the same range endpoint $R$ and bit width $b=3$. Red regions mark the zero bins. (a)~Uniform quantization has $d=s$. (b)~Magnitude pruning followed by uniform quantization (P+U) makes the dashed levels unreachable. (c)~CoDeQ widens the zero bin while redistributing all nonzero levels over the retained range.}
\label{fig:quantizer_comparison}
\end{minipage}
\par\medskip

\paragraph{Reconstruction and weight sparsity.}
The quantizer jointly selects which weights become zero and which quantized values represent the retained weights. For a fixed codebook, the following proposition shows that these assignments minimize a common objective: reconstruction error plus a term proportional to the number of nonzero weights.

\begin{proposition}[Reconstruction and nonzero cost]
\label{prop:assignment}
\label{prop:variational}
For fixed $d\ge0$, $s>0$, and integer $b\ge2$, let $\rho=(d^2-s^2)/8$ and let $\mathcal C$ denote the codebook (set of scalar reconstruction levels) of the dead-zone quantizer in \Cref{eq:operator}. Then
\begin{equation}
 \tilde q(\w;s,d,b)\in\operatorname*{arg\,min}_{\mathbf z\in\mathcal C^{\,n}}
 \left\{\tfrac12\|\w-\mathbf z\|_2^2+\rho\|\mathbf z\|_0\right\},
\label{eq:assignment}
\end{equation}
with selection among tied minimizers at decision boundaries.
\end{proposition} 
The proof is in Appendix~\ref{app:variational_proof}. For $d>s$, a nonzero assignment must reduce reconstruction error by at least $\rho$ to be preferred over zero. At $d=s$, this cost vanishes; for $d<s$, the negative term favors nonzero assignments. Thus the zero-bin width determines the reconstruction--sparsity trade-off for a given codebook. We next choose the spacing $s$ of its nonzero levels.

\paragraph{Allocating the nonzero levels.}
Once the zero-bin width and bit width are fixed, we choose the $Q_b$ positive reconstruction levels to minimize worst-case error over the retained range. We keep the largest level at $R$ so that moving the pruning threshold leaves the saturation magnitude unchanged. The following proposition gives the optimal allocation under this constraint.

\begin{proposition}[Optimal spacing of the nonzero levels]
\label{prop:minimax}
Fix $0\le d/2<R$ and at most $Q_b=2^{b-1}-1$ positive reconstruction levels, with the largest level fixed at $R$. Let \(\mathcal A\) denote a candidate set of positive reconstruction levels. The optimal worst-case error for retained positive inputs is
\begin{equation}
\begin{aligned}
 E_*&=\min_{\substack{\mathcal A\subset(0,R]\\ |\mathcal A|\le Q_b\\R\in\mathcal A}}
 \sup_{t\in(d/2,R]}\min_{r\in\mathcal A}|t-r|=\frac{R-d/2}{2Q_b-1}.
\end{aligned}
\label{eq:minimax}
\end{equation}
\end{proposition}
The proof is in Appendix~\ref{app:minimax_proof}. Between adjacent reconstruction levels, the largest nearest-level error occurs at their midpoint and equals half their spacing. The optimal construction therefore uses the step size
\begin{equation}
 s=2E_*=\frac{2R-d}{2Q_b-1}.
\label{eq:scale}
\end{equation}
With this choice, the CoDeQ codebook in \Cref{eq:operator} attains the bound, with maximum reconstruction error $s/2$ on $(d/2,R]$.
At fixed range and bit width, widening the zero bin prunes more weights while concentrating the available nonzero levels over a smaller interval. This reduces the worst-case reconstruction error for retained inputs, while the additional pruned weights contribute pruning error. The bound excludes inputs outside the chosen range.
In P+U, the pruning mask can make inner nonzero levels of the unchanged uniform grid unreachable (\Cref{fig:quantizer_comparison}(b)).

The ablation in \Cref{tab:prune_uniform_w4} compares CoDeQ with P+U on ResNet-20/CIFAR-10 at W4/A32. CoDeQ keeps all 14 signed nonzero levels reachable and gives higher accuracy and lower MSE at the three selected points.

\begin{table}[!t]
\centering
\footnotesize
\setlength{\tabcolsep}{2.25pt}
\renewcommand{\arraystretch}{1.05}
\caption{CoDeQ versus learnable magnitude pruning followed by uniform quantization (P+U) on ResNet-20/CIFAR-10 at W4/A32 over 100 epochs. CoDeQ achieves higher test accuracy and lower reconstruction error across the three evaluated compression settings, while keeping all reconstruction levels available.}
\label{tab:prune_uniform_w4}
\input{files/table_prune_uniform_w4}
\par\vspace{3pt}
\begin{minipage}{\columnwidth}
\scriptsize\raggedright
MSE compares nonzero quantized weights with clipped latent weights. Levels are reachable signed nonzero values (14 for 4-bit), averaged equally across layers.
\end{minipage}
\end{table}

\subsection{Joint compression objective}
\label{sec:objective}
\paragraph{Parameterization.}
CoDeQ supports fixed-bit and MP training. Both learn the latent weights and layerwise pruning thresholds; MP also learns the layerwise bit widths, while fixed-bit training uses a prescribed bit-width $b_0$. For each layer, the retained-range fraction $\theta_{\mathrm{dz}}\in[0,1]$ measures the fraction of $[0,R]$ above the pruning threshold and determines the zero-bin width:
\begin{equation}
 d(\theta_{\mathrm{dz}})=2R(1-\theta_{\mathrm{dz}}).
\label{eq:range_coordinate}
\end{equation}
This geometric fraction need not equal the measured nonzero fraction $p_\ell$ of quantized weights. Reducing $\theta_{\mathrm{dz}}$ widens the zero bin and reduces the spacing between nonzero levels. For the learned bit widths, we introduce a normalized coordinate $\theta_{\mathrm{bit}}$ and a continuous, unrounded bit width $\beta$:
\begin{equation}
\begin{aligned}
 \theta_{\mathrm{bit}}&\in[b_{\min}/b_{\max},1],\\
 \beta&=b_{\max}\theta_{\mathrm{bit}},\\
 b(\theta_{\mathrm{bit}})&=\lfloor\beta\rceil.
\end{aligned}
\label{eq:bit_coordinate}
\end{equation}
Normalizing by $b_{\max}$ places the learned precision coordinate on a dimensionless scale comparable to $\theta_{\mathrm{dz}}$.
We abbreviate $d=d(\theta_{\mathrm{dz}})$ and $b=b(\theta_{\mathrm{bit}})$.

\paragraph{Compression objective.}
Let $\mathcal L_{\rm task}(\hat W)$ denote the task loss evaluated with the quantized weights. We use the regularized loss with compression penalty $\lambda\mathcal R$:
\begin{equation}
 \mathcal L=\mathcal L_{\rm task}(\hat W)
           +\lambda\mathcal R,
\label{eq:objective}
\end{equation}
with $\lambda\ge0$ and
\begin{equation}
 \mathcal R=\sum_\ell c_\ell b(\theta_{\mathrm{bit},\ell})\,\theta_{\mathrm{dz},\ell},
\label{eq:regularizer}
\end{equation}
where $c_\ell=\frac{\mathrm{MACs}_\ell\,a_\ell}{32^2\sum_j \mathrm{MACs}_j}$ weights each layer by its dense MAC count and fixed activation precision $a_\ell$, normalized by the dense FP32 cost. The sums cover the same weighted layers as the BOP metric. For fixed-bit training, $b(\theta_{\mathrm{bit},\ell})$ is replaced by $b_0$.

Reducing either factor in $b\theta_{\mathrm{dz}}$ encourages compression: lower precision or a wider zero bin. At fixed weights and precision, the measured nonzero fraction $p_\ell$ is piecewise constant in the pruning threshold and provides no ordinary gradient between crossings. We therefore use $\theta_{\mathrm{dz}}=1-d/(2R)$ as a geometric proxy to supply a direct compression gradient.

\Cref{fig:compression_behavior}\subref{fig:lambda_sensitivity_w4} illustrates the effect of compression pressure at fixed W4 on ResNet-20/CIFAR-10. Increasing $\lambda$ from 1 to 18 reduces mean relative BOPs from 10.34\% to 1.38\%, while mean test accuracy changes from 90.71\% to 87.14\%.

\subsection{Learning the quantizer}
\label{sec:backward}
\label{sec:codeq}
We optimize \Cref{eq:objective} using surrogate gradients through the hard quantizer. During training, we recompute $R$ as the 99th percentile of the latent-weight magnitudes and hold it fixed during backpropagation:
\begin{equation}
 R=\operatorname{stopgrad}\!\left(\operatorname{P99}(|\w|)\right).
\label{eq:training_range}
\end{equation}
The retained-range fraction in \Cref{eq:range_coordinate} enforces $0\le d/2\le R$. To keep the step size positive at $d/2=R$, we add a numerical stabilizer to the spacing in \Cref{eq:scale}:
\begin{equation}
 s(d,b)=\frac{R-d/2}{Q_b-\tfrac12}+\epsilon,\qquad\epsilon=10^{-8}.
\label{eq:training_scale}
\end{equation}
We initialize both $\theta_{\mathrm{dz}}=0.995$ and, for MP, $\theta_{\mathrm{bit}}=0.995$, giving a narrow zero bin and an initial bit width of $b=8$.
Algorithm~\ref{alg:training} summarizes the full training procedure. After the optimizer proposes $\theta_{\mathrm{dz}}',\theta_{\mathrm{bit}}'$, we project outside autograd:
\begin{equation}
 \theta_{\mathrm{dz}}^+=\Pi_{[0,1]}(\theta_{\mathrm{dz}}'),\qquad
 \theta_{\mathrm{bit}}^+=\Pi_{[b_{\min}/b_{\max},1]}(\theta_{\mathrm{bit}}').
\label{eq:projection}
\end{equation}

\begin{algorithm}[t]
\small
\caption{CoDeQ training}
\label{alg:training}
\label{alg:codeq_dz}
\KwIn{Data $\mathcal D$, initialized $W,\theta_{\mathrm{dz}}$, optional $\theta_{\mathrm{bit}}$, coefficient $\lambda$, optimizer \textnormal{Opt}}
\ForEach{training minibatch}{
 \ForEach{weighted layer $\ell$}{
  Recompute detached $R_\ell$ by \Cref{eq:training_range}\;
  Set $d_\ell,b_\ell$ by \Cref{eq:range_coordinate,eq:bit_coordinate}; use $b_0$ for fixed-bit training\;
  Set $s_\ell$ by \Cref{eq:training_scale} and quantize with \Cref{eq:operator}\;
  For MP, form the detached adjacent-grid difference in \Cref{eq:adjacent}\;
 }
 Evaluate $\mathcal L_{\rm task}(\hat W)$ and $\mathcal R$\;
 Update using \Cref{eq:combined_gradients} and \textnormal{Opt}\;
 Project compression parameters by \Cref{eq:projection}\;
}
\end{algorithm}

\paragraph{Weight gradients.}
Let $g_i=\partial\mathcal L_{\rm task}/\partial\hat w_i$ denote the task gradient with respect to a quantized weight. Rounding blocks the encoder's ordinary gradients, while standard ReLU and clipping derivatives can block updates that would let pruned or clipped weights recover. For the operator in \Cref{eq:operator}, we therefore use identity STEs through rounding, ReLU, and the clipping input, preserving the hard forward quantizer. Both sign operations have zero backward derivatives, and clipping bounds receive no gradients. This gives the task gradient
$
 \widetilde\partial_{w_i}\mathcal L_{\rm task}
   =g_i\mathbf1_{\{w_i\ne0\}},
$ derived in Appendix~\ref{app:backward_derivatives}.
Nonzero latent weights therefore receive feedback even when they quantize to zero or a clipping level. Exactly zero latent weights remain an exception under the standard derivative of absolute value.

\paragraph{Gradients for the retained-range fraction.}
At fixed $R$ and selected bit width $b$, changing $\theta_{\mathrm{dz}}$ changes both the pruning threshold and the step size between nonzero levels. Using the same STE rules, let $J_{\mathrm{dz},i}=\widetilde\partial_{\theta_{\mathrm{dz}}}\hat w_i$ denote the quantizer's surrogate derivative, including the threshold, step size, encoder, and decoder paths. The task gradient is
$
 \widetilde\partial_{\theta_{\mathrm{dz}}}\mathcal L_{\rm task}
   =\sum_i g_iJ_{\mathrm{dz},i}.
$
Appendix~\ref{app:backward_derivatives} derives these surrogate derivatives and gives the closed form of $J_{\mathrm{dz},i}$.

\paragraph{Bit-width gradients.}
In MP training, we need a task gradient for the learned bit width.
We estimate it by quantizing the same weights at two adjacent integer
bit widths and measuring how their quantized values change.
For the unrounded bit width $\beta=b_{\max}\theta_{\mathrm{bit}}$,
we define
\begin{equation}
\begin{aligned}
 b^-&=\clip(\lfloor\beta\rfloor,b_{\min},b_{\max}-1),\\
 \Delta_i&=\sg\big[\tilde q(w_i;s(d,b^-+1),d,b^-+1)\\
 &\hspace{14mm}-\tilde q(w_i;s(d,b^-),d,b^-)\big].
\end{aligned}
\label{eq:adjacent}
\end{equation}
Both quantizers use the same latent weights $\w$, range $R$, and
pruning threshold $d$. Thus $\Delta_i$ measures the change in
quantized weight $i$ when precision increases by one bit.
We use this difference as a surrogate derivative with respect to
$\beta$, treating $\Delta_i$ as constant during backpropagation.

Combining these changes with the task gradients $g_i$ gives
$
 \widetilde\partial_{\theta_{\mathrm{bit}}}\mathcal L_{\rm task}
   =b_{\max}\sum_i g_i\Delta_i.
$
The sum estimates the resulting change in task loss to first order:
a negative value favors increasing precision. A weight contributes
zero when its quantized value is identical at both bit widths.

%The forward pass uses the selected bit width $b=\lfloor\beta\rceil$.
%Its task path is detached from $\beta$, while retaining the weight
%and retained-range surrogate derivatives described above. 

\Cref{fig:adjacent_grid_stability} shows smaller fluctuations in test accuracy and fewer layerwise bit-width changes with adjacent-grid feedback on ResNet-20/CIFAR-10.

\begin{figure}[!t]
\centering
\includegraphics[width=0.9\columnwidth]{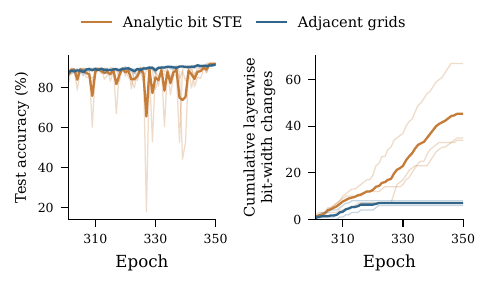}
\vspace{-0.25cm}
\caption{Comparison of bit-width gradient estimators on ResNet-20/CIFAR-10, MP/A32. Adjacent grid gradients reduces accuracy oscillations and bit-width changes in late training. Left: test accuracy. Right: cumulative layerwise bit-width changes over epochs 301--350. Thin lines show individual seeds; thick lines show three-seed means.}
\vspace{-0.25cm}
\label{fig:adjacent_grid_stability}
\end{figure}

\paragraph{Normalization and a shared learning rate.}
Combining task and compression gradients gives
\begin{equation}
\begin{aligned}
 \widetilde\partial_{w_i}\mathcal L
   &=g_i\mathbf1_{\{w_i\ne0\}},\\
 \widetilde\partial_{\theta_{\mathrm{dz}}}\mathcal L
   &=\sum_i g_iJ_{\mathrm{dz},i}+\lambda c b,\\
 \widetilde\partial_{\theta_{\mathrm{bit}}}\mathcal L
   &=b_{\max}\!\left[\sum_i g_i\Delta_i
                         +\lambda c \theta_{\mathrm{dz}}\right].
\end{aligned}
\label{eq:combined_gradients}
\end{equation}
The nonnegative compression gradients push the two compression parameters downward under gradient descent.
We first compare the penalty gradients for the two compression parameters. Learning the unrounded bit width $\beta$ directly would give a compression gradient $\lambda c\theta_{\mathrm{dz}}$. With the normalized parameter and an identity STE for bit rounding, the compression gradients are
\begin{equation}
\begin{aligned}
 \partial_{\theta_{\mathrm{dz}}}(\lambda\mathcal R)
   &=\lambda c b,\\
 \widetilde\partial_{\theta_{\mathrm{bit}}}(\lambda\mathcal R)
   &=\lambda c b_{\max}\theta_{\mathrm{dz}}.
\end{aligned}
\label{eq:compression_coordinate_gradients}
\end{equation}
Near $\theta_{\mathrm{dz}}\approx1$ and $b\approx b_{\max}$, both gradients are approximately $\lambda c b_{\max}$. Their matching initial scales motivate a shared learning rate $\eta$ for the two compression parameters, alongside the single compression coefficient $\lambda$. %For plain SGD learning $\theta_{\mathrm{bit}}$ at rate $\eta$ is equivalent to learning the unrounded bit width $\beta$ at rate $\eta b_{\max}^{2}$. Appendix~\ref{app:precision_update_scaling} gives the derivation.
%\paragraph{Combining task and compression feedback.}
%The compression penalty has no direct latent-weight gradient. Weight decay is zero for compression parameters; model-weight decay is a separate optimizer setting. Fixed-bit training omits the final line.

\Cref{fig:compression_behavior}\subref{fig:joint_compression_trajectory}
illustrates how precision and sparsity evolve from the prescribed initialization, with $\theta_{\mathrm{dz}}$ and $\theta_{\mathrm{bit}}$ using the same compression learning rate. On SimpleViT/CIFAR-10, the mean rounded bit width decreases from 8 to 3.1 bits over 100 epochs, while weight sparsity grows to 74\%.

\begin{figure}[!t]
\centering
\begin{subfigure}[b]{0.45\columnwidth}
\centering
\includegraphics[width=\linewidth]{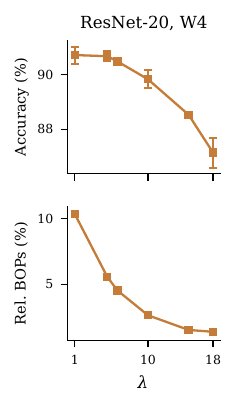}
\caption{Compression pressure.}
\label{fig:lambda_sensitivity_w4}
\end{subfigure}
\hfill
\begin{subfigure}[b]{0.45\columnwidth}
\centering
\includegraphics[width=\linewidth]{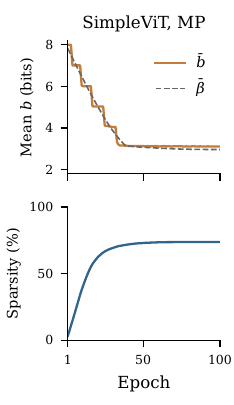}
\caption{Joint adaptation.}
\label{fig:joint_compression_trajectory}
\end{subfigure}
\vspace{-0.25cm}
\caption{Compression behavior on CIFAR-10. (a)~ResNet-20, W4/A32: final-epoch accuracy and relative BOPs (mean $\pm$ sample SD over three seeds), varying only $\lambda$. (b)~SimpleViT, MP/A32: mean rounded $b$ and continuous $\beta$ and total weight sparsity over 100 epochs with a shared compression learning rate. We see a stable convergence of both bit-widths and sparsity.}
\label{fig:compression_behavior}
\vspace{-0.25cm}
\end{figure}

\endgroup

%% file: files/table_prune_uniform_w4.tex
% Generated from plots/data/prune_uniform_w4.csv; uses the paper's shared table macros.
\cqtabular{prune_uniform_w4}{rrrr}{%
\toprule
Acc.\,(\%)\,$\uparrow$ & Rel.\ BOPs\,(\%)\,$\downarrow$ & MSE\,$\times 10^3$\,$\downarrow$ & Levels\,$\uparrow$\\
\midrule
\multicolumn{4}{@{}l}{\textbf{P+U}}\\
\cqresult{89.22}{0.29} & \cqresult{4.91}{0.10} & \cqresult{0.72}{0.01} & \cqresult{14.00}{0.00} \\
\cqresult{88.27}{0.29} & \cqresult{2.67}{0.17} & \cqresult{0.89}{0.04} & \cqresult{12.40}{0.00} \\
\cqresult{86.33}{0.53} & \cqresult{1.59}{0.10} & \cqresult{1.06}{0.02} & \cqresult{8.77}{0.12} \\
\arrayrulecolor{black!40}\cmidrule[0.25pt](l{1em}r{1em}){1-4}\arrayrulecolor{black}
\multicolumn{4}{@{}l}{\textbf{CoDeQ}}\\
\cqresult{90.48}{0.12} & \cqresult{4.52}{0.01} & \cqresult{0.47}{0.01} & \cqresult{14.00}{0.00} \\
\cqresult{89.83}{0.33} & \cqresult{2.64}{0.01} & \cqresult{0.52}{0.07} & \cqresult{14.00}{0.00} \\
\cqresult{88.53}{0.03} & \cqresult{1.52}{0.02} & \cqresult{0.66}{0.03} & \cqresult{14.00}{0.00} \\
\bottomrule
}

%% file: files/experiments.tex
\section{Experiments}
\label{sec:experiments}
\paragraph{Evaluation.}
We evaluate CoDeQ on image classification with convolutional networks (ResNet-20~\cite{he2016resnet} and VGG7~\cite{simonyan2015vgg}) and vision transformers (SimpleViT~\cite{beyer2022simplevit,wang2020vitpytorch}, DeiT-Tiny~\cite{touvron2021deit}, and ViT-Small~\cite{dosovitskiy2021vit}), and on question answering with BERT-base-uncased~\cite{devlin2019bert}. We compare fixed 4-bit weights (W4) and learned layerwise mixed precision (MP) with published joint pruning--quantization results. Performance is measured by top-1 accuracy on CIFAR-10~\cite{krizhevsky2009cifar} and ImageNet-1K~\cite{deng2009imagenet,russakovsky2015ilsvrc}, and exact match (EM) and token-level F1 on SQuAD 1.1~\cite{rajpurkar2016squad}. CoDeQ results are final-epoch means $\pm$ sample standard deviations over three independent seeds. Training and evaluation settings are given in Appendix~\ref{app:experimental_setup}.

\subsection{Computer vision}
\label{sec:vision_experiments}
\subsubsection{Convolutional networks}
\textbf{ResNet-20 on CIFAR-10.}
\Cref{tab:cnn_results} compares CoDeQ with SQL, QST-Q, and GETA using full-precision activations. W4 achieves $92.03\%$ accuracy at $2.47\%$ relative BOPs, improving on all three methods in both accuracy and cost. Compared with QST-Q, it gains $0.43$ pp while using approximately $25\%$ fewer BOPs. MP reduces relative BOPs further to $1.36\%$, retaining $91.55\%$ accuracy, only $0.15$ pp below the FP32 reference.\\
\textbf{VGG7 on CIFAR-10.}
We use 2-bit activations with an 8-bit image input. W4 and MP reach $92.25\%$ and $92.73\%$ accuracy at $0.06\%$ and $0.05\%$ relative BOPs, respectively. Both improve on DJPQ in accuracy and cost, and MP also improves on GETA. Bayesian Bits (BB) and FITCompress retain higher accuracy at higher cost; compared with FITCompress, MP trades $0.34$ pp accuracy for approximately $4.5\times$ lower BOPs.

\begin{table}[!t]
\centering
\footnotesize
\setlength{\tabcolsep}{2.25pt}
\renewcommand{\arraystretch}{1.05}
\caption{CNN comparisons with published compression methods. Model-name rows report FP32 accuracy (\%); method rows report differences from that reference (pp).}
\label{tab:cnn_results}
\cqtabular{cnn}{lcc!{\hspace{4pt}}r!{\hspace{-4pt}}r}{%
\toprule
Model & P & W/A & $\Delta$Acc.\,(pp)\,$\uparrow$ & Rel.\ BOPs\,(\%)\,$\downarrow$\\
\midrule
\multicolumn{5}{@{}l}{\textit{CIFAR-10}}\\
\addlinespace[1.5pt]
\underline{ResNet-20} & \textemdash & FP32 & \cqresult{91.70}{} & \cqresult{100.00}{}\\
SQL~\cite{joint:SQL} & U & $\mathrm{MP}/32$ & \cqresult{-0.80}{} & \cqresult{6.10}{}\\
QST-Q~\cite{joint:QST} & U & $\mathrm{MP}/32$ & \cqresult{-0.10}{} & \cqresult{3.30}{}\\
GETA~\cite{joint:qu2025automatic} & S & $\mathrm{MP}/32$ & \cqresult{-0.28}{} & \cqresult{4.50}{}\\
\arrayrulecolor{black!40}\cmidrule[0.25pt](l{1em}r{1em}){1-5}\arrayrulecolor{black}
\textbf{CoDeQ} & U & $4/32$ & \cqresult{+0.33}{0.23} & \cqresult{2.47}{0.16}\\
\textbf{CoDeQ} & U & $\mathrm{MP}/32$ & \cqresult{-0.15}{0.09} & \cqresult{1.36}{0.02}\\
\midrule
\underline{VGG7} & \textemdash & FP32 & \cqresult{93.05}{} & \cqresult{100.00}{}\\
DJPQ~\cite{joint:wang2020djpq} & S & $\mathrm{MP}/\mathrm{MP}$ & \cqresult{-1.51}{} & \cqresult{0.48}{}\\
BB~\cite{joint:van2020bayesian} & S & $\mathrm{MP}/\mathrm{MP}$ & \cqresult{+0.18}{} & \cqresult{0.51}{}\\
GETA~\cite{joint:qu2025automatic} & S & $\mathrm{MP}/\mathrm{MP}$ & \cqresult{-0.48}{} & \cqresult{0.41}{}\\
FITCompress~\cite{joint:FITCompress} & U & $\mathrm{MP}/\mathrm{MP}$ & \cqresult{+0.02}{} & \cqresult{0.23}{}\\
\arrayrulecolor{black!40}\cmidrule[0.25pt](l{1em}r{1em}){1-5}\arrayrulecolor{black}
\textbf{CoDeQ} & U & $4/2$ & \cqresult{-0.80}{0.15} & \cqresult{0.06}{0.00}\\
\textbf{CoDeQ} & U & $\mathrm{MP}/2$ & \cqresult{-0.32}{0.18} & \cqresult{0.05}{0.00}\\
%\midrule
%\multicolumn{5}{@{}l}{\textit{ImageNet-1K}}\\
%\addlinespace[1.5pt]
% MobileNetV2 references: torchvision V1 documented top-1 = 71.878%;
% https://docs.pytorch.org/vision/stable/models/generated/torchvision.models.mobilenet_v2.html
% BB: arXiv:2005.07093v3 source figures/mobilenet-v2.pdf, Figure 2(b).
% Selected (absolute top-1, relative BOPs): highest accuracy
% (71.9566667,6.2566667), and lowest BOPs (63.28,1.1266667).
% DJPQ Table 3: top-1 = 69.30%, dense/compressed BOPs = 5682.32/132.28 G;
% relative BOPs = 100*132.28/5682.32 = 2.3279224%. Cost not harmonized.
% Figure-based reconstruction (Appendix Figs. 11--12) fails the source-total
% check: 5.1159555 G MACs / 150.0193043 G BOPs vs. Table 3's 4.76 / 132.28.
% Keep the published source-native percentage; do not substitute the unverified recount.
% Accuracy deltas below subtract the exact common reference 71.878 before rounding.
%\textbf{MobileNetV2} & \textemdash & FP32 & \cqresult{71.88}{} & \cqresult{100.00}{}\\
%BB$^\dagger$~\cite{joint:van2020bayesian} & S & $\mathrm{MP}/\mathrm{MP}$ & \cqresult{+0.08}{} & \cqresult{6.26}{}\\
%BB$^\dagger$ & S & $\mathrm{MP}/\mathrm{MP}$ & \cqresult{-8.60}{} & \cqresult{1.13}{}\\
%DJPQ~\cite{joint:wang2020djpq} & S & $\mathrm{MP}/\mathrm{MP}$ & \cqresult{-2.58}{} & \cqresult{2.33}{}\\
%\arrayrulecolor{black!40}\cmidrule[0.25pt](l{1em}r{1em}){1-5}\arrayrulecolor{black}
%\textbf{CoDeQ} & U & $4/4$ & \cqresult{\text{\textemdash}}{} & \cqresult{\text{\textemdash}}{}\\
%\textbf{CoDeQ} & U & $\mathrm{MP}/4$ & \cqresult{\text{\textemdash}}{} & \cqresult{\text{\textemdash}}{}\\
\bottomrule
}
\par\vspace{3pt}
\begin{minipage}{\columnwidth}
\scriptsize\raggedright
P: S = structured, U = unstructured. W/A: weight/activation bits; MP = mixed precision.
\end{minipage}
% Previously commented comparison rows, retained for reference:
 %& QST-P~\cite{joint:QST} & Unstructured & MP W/A32 & $+0.10$ & $5.00$\\
 %& QST-B~\cite{joint:QST} & Unstructured & MP W/A32 & $-0.20$ & $5.10$\\
 %& DJPQ-restrict~\cite{joint:wang2020djpq} & Structured & MP W/A & $-1.62$ & $0.46$\\
 %& Bayesian Bits ($\mu{=}0.1$)~\cite{joint:van2020bayesian} & Structured & MP W/A & $-1.09\!\pm\!0.04$ & $0.29\!\pm\!0.00$\\
 %& FITCompress$^{\dagger}$~\cite{joint:FITCompress} & Unstructured & MP W/A & $-1.03\!\pm\!0.02$ & $0.29\!\pm\!0.01$\\
 %& DJPQ~\cite{joint:wang2020djpq} & Structured & MP W/A & $-1.03$ & $1.89$\\
 %& DJPQ~\cite{joint:wang2020djpq} & Structured & MP W/A & $-1.50$ & $1.67$\\
 %& DJPQ-restrict~\cite{joint:wang2020djpq} & Structured & MP W/A & $-1.18$ & $1.91$\\
 %& BB-updated ($\mu{=}0.01$)~\cite{joint:van2020bayesian} & Structured & MP W/A & $-0.48\!\pm\!0.07$ & $2.41\!\pm\!0.03$\\
 %& BB-updated ($\mu{=}0.03$)~\cite{joint:van2020bayesian} & Structured & MP W/A & $-1.12\!\pm\!0.08$ & $1.70\!\pm\!0.03$\\
 %& BB-updated ($\mu{=}0.05$)~\cite{joint:van2020bayesian} & Structured & MP W/A & $-2.13\!\pm\!0.18$ & $1.38\!\pm\!0.03$\\
 %& FITCompress~\cite{joint:FITCompress} & Unstructured & MP W/A & $-1.46\!\pm\!0.12$ & $1.38\!\pm\!0.06$\\
 \vspace{-0.5cm}
\end{table}

\subsubsection{Vision transformers}
\textbf{SimpleViT on CIFAR-10.}
\Cref{tab:transformer_results} shows that W4 reaches $86.33\%$ accuracy at $3.14\%$ relative BOPs, while MP improves mean accuracy to $86.81\%$ and lowers cost to $2.63\%$. MP exceeds the FP32 reference by $0.33$ pp. Both settings retain full-precision activations and give higher mean accuracy at lower BOPs than GETA.\\
\textbf{DeiT-Tiny on ImageNet-1K.}
W4 and MP both exceed the FP32 reference, reaching $73.26\%$ and $72.38\%$ accuracy at $9.03\%$ and $8.12\%$ relative BOPs, respectively. W4 gives the highest accuracy among the compared methods and also reduces BOPs relative to GETA. MP offers additional compression at a lower accuracy.\\
\textbf{ViT-Small on ImageNet-1K.}
W4 reaches $82.13\%$ accuracy at $8.55\%$ relative BOPs, exceeding FP32 by $0.75$ pp. MP lowers cost to $6.83\%$ at $81.26\%$ accuracy, $0.12$ pp below FP32. Both settings achieve higher mean accuracy at lower BOPs than GETA.
\begin{table}[!t]
\centering
\footnotesize
\setlength{\tabcolsep}{2.25pt}
\renewcommand{\arraystretch}{1.05}
\caption{Vision-transformer comparisons with GETA. Model-name rows report FP32 accuracy (\%); method rows report differences from that reference (pp).}
\label{tab:transformer_results}
\cqtabular{vit}{lcc!{\hspace{4pt}}r!{\hspace{-4pt}}r}{%
\toprule
Model & P & W/A & $\Delta$Acc.\,(pp)\,$\uparrow$ & Rel.\ BOPs\,(\%)\,$\downarrow$\\
\midrule
\multicolumn{5}{@{}l}{\textit{CIFAR-10}}\\
\addlinespace[1.5pt]
\underline{SimpleViT} & \textemdash & FP32 & \cqresult{86.48}{} & \cqresult{100.00}{}\\
GETA~\cite{joint:qu2025automatic} & S & $\mathrm{MP}/32$ & \cqresult{-0.42}{} & \cqresult{4.95}{}\\
\arrayrulecolor{black!40}\cmidrule[0.25pt](l{1em}r{1em}){1-5}\arrayrulecolor{black}
\textbf{CoDeQ} & U & $4/32$ & \cqresult{-0.15}{0.68} & \cqresult{3.14}{0.05}\\
\textbf{CoDeQ} & U & $\mathrm{MP}/32$ & \cqresult{+0.33}{1.18} & \cqresult{2.63}{0.07}\\
\midrule
\multicolumn{5}{@{}l}{\textit{ImageNet-1K}}\\
\addlinespace[1.5pt]
\underline{DeiT-Tiny} & \textemdash & FP32 & \cqresult{72.16}{} & \cqresult{100.00}{}\\
GETA~\cite{joint:qu2025automatic} & S & $\mathrm{MP}/32$ & \cqresult{+0.72}{} & \cqresult{16.95}{}\\
\arrayrulecolor{black!40}\cmidrule[0.25pt](l{1em}r{1em}){1-5}\arrayrulecolor{black}
\textbf{CoDeQ} & U & $4/32$ & \cqresult{+1.10}{0.06} & \cqresult{9.03}{0.00}\\
\textbf{CoDeQ} & U & $\mathrm{MP}/32$ & \cqresult{+0.22}{0.03} & \cqresult{8.12}{0.10}\\
\midrule
\underline{ViT-Small} & \textemdash & FP32 & \cqresult{81.38}{} & \cqresult{100.00}{}\\
GETA~\cite{joint:qu2025automatic} & S & $\mathrm{MP}/32$ & \cqresult{-1.26}{} & \cqresult{19.37}{}\\
\arrayrulecolor{black!40}\cmidrule[0.25pt](l{1em}r{1em}){1-5}\arrayrulecolor{black}
\textbf{CoDeQ} & U & $4/32$ & \cqresult{+0.75}{0.05} & \cqresult{8.55}{0.01}\\
\textbf{CoDeQ} & U & $\mathrm{MP}/32$ & \cqresult{-0.12}{0.10} & \cqresult{6.83}{0.07}\\
%\midrule
%\textbf{Swin-Tiny} & \textemdash & FP32 & \cqresult{81.38}{} & \cqresult{100.00}{}\\
%GETA~\cite{joint:qu2025automatic} & S & $\mathrm{MP}/32$ & \cqresult{-1.29}{} & \cqresult{21.84}{}\\
%\arrayrulecolor{black!40}\cmidrule[0.25pt](l{1em}r{1em}){1-5}\arrayrulecolor{black}
%\textbf{CoDeQ} & U & $4/32$ & \cqresult{\text{\textemdash}}{} & \cqresult{\text{\textemdash}}{}\\
%\textbf{CoDeQ} & U & $\mathrm{MP}/32$ & \cqresult{\text{\textemdash}}{} & \cqresult{\text{\textemdash}}{}\\
%\midrule
%\textbf{Swin-Base} & \textemdash & FP32 & \cqresult{83.60}{} & %\cqresult{100.00}{}\\
%\textbf{CoDeQ} & U & $4/32$ & \cqresult{\text{\textemdash}}{} & %\cqresult{\text{\textemdash}}{}\\
%\textbf{CoDeQ} & U & $\mathrm{MP}/32$ & %\cqresult{\text{\textemdash}}{} & \cqresult{\text{\textemdash}}%{}\\
\bottomrule
}
\end{table}

\begin{figure}[!t]
\centering
\includegraphics[width=\columnwidth]{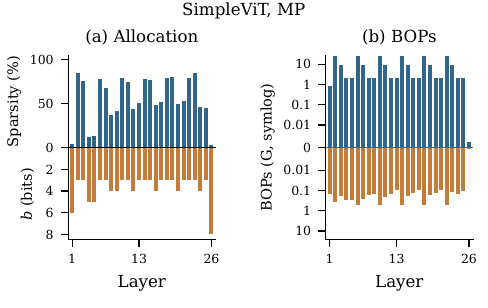}
\vspace{-0.75cm}
\caption{Learned layerwise allocation on SimpleViT/CIFAR-10, MP/A32, averaged over three seeds. (a)~Weight sparsity (up) and precision $b_\ell$ (down). We see low sparsity and high bit-widths allocated for the input and output layer. (b)~Dense FP32 (up) and CoDeQ (down) costs share one symmetric-log axis in G BOPs. Layers with high BOPs values are compressed more, as expected from the objective.}
\label{fig:learned_layerwise_allocation}
\vspace{-0.75cm}
\end{figure}

\flushleft
\textbf{Learned layerwise allocation.}
\Cref{fig:learned_layerwise_allocation} shows how SimpleViT distributes sparsity and precision under one global compression coefficient. Attention projections use 3-bit weights with high sparsity, while the patch embedding and classifier retain 6-bit and 8-bit weights, respectively, in all three runs. The resulting allocation concentrates compression in the costly attention projections while preserving precision in the input and output layers.

\subsection{Language models}
\label{sec:language_experiments}
% \paragraph{BERT question answering.}
\Cref{tab:language_results} reports BERT-base-uncased on SQuAD 1.1 with full-precision activations. W4 reaches $80.09\%$ EM and $87.87\%$ F1 at $8.45\%$ relative BOPs, nearly matching the FP32 EM reference with a $0.63$ pp F1 decrease. MP reaches $79.82\%$ EM and $87.64\%$ F1 at $9.47\%$ relative BOPs. Both settings achieve higher EM and F1 at lower BOPs than GETA. W4 gives the strongest accuracy--BOP trade-off in this comparison.

\begin{table}[!t]
\centering
\footnotesize
\setlength{\tabcolsep}{2pt}
\renewcommand{\arraystretch}{1.05}
\caption{BERT question answering on SQuAD. The model-name row reports absolute FP32 EM and F1 (\%); method rows report differences from that reference (pp).}
\vspace{-0.25cm}
\label{tab:language_results}
\cqtabular{language}{lccrrr}{%
\toprule
Model & P & W/A & $\Delta$EM\,(pp)\,$\uparrow$ & $\Delta$F1\,(pp)\,$\uparrow$ & Rel.\ BOPs\,(\%)\,$\downarrow$\\
\midrule
\multicolumn{6}{@{}l}{\textit{SQuAD}}\\
\addlinespace[1.5pt]
\underline{BERT} & \textemdash & FP32 & \cqresult{80.08}{} & \cqresult{88.50}{} & \cqresult{100.00}{}\\
% GETA Appendix C/Table 7: learned weight bits in [4,16], initialized at 8.
% Released BERT setup uses the default WEIGHT_ONLY mode:
% https://github.com/microsoft/geta/blob/0518a57021eeed0b6f74eab709730dcf5b0135f1/sanity_check/test_qbert.py#L29
% https://github.com/microsoft/geta/blob/0518a57021eeed0b6f74eab709730dcf5b0135f1/only_train_once/quantization/quant_layers.py#L397-L399
GETA~\cite{joint:qu2025automatic} & S & $\mathrm{MP}/32$ & \cqresult{-1.82}{} & \cqresult{-2.44}{} & \cqresult{19.40}{}\\
\arrayrulecolor{black!40}\cmidrule[0.25pt](l{1em}r{1em}){1-6}\arrayrulecolor{black}
\textbf{CoDeQ} & U & $4/32$ & \cqresult{+0.01}{0.18} & \cqresult{-0.63}{0.13} & \cqresult{8.45}{0.03}\\
\textbf{CoDeQ} & U & $\mathrm{MP}/32$ & \cqresult{-0.26}{0.19} & \cqresult{-0.86}{0.20} & \cqresult{9.47}{0.03}\\
\bottomrule
}
\vspace{-0.25cm}
\end{table}

\subsection{Kernel execution}
\label{sec:kernel_experiments}

\begin{figure}[t]
    \centering
    \includegraphics[width=0.8\columnwidth]{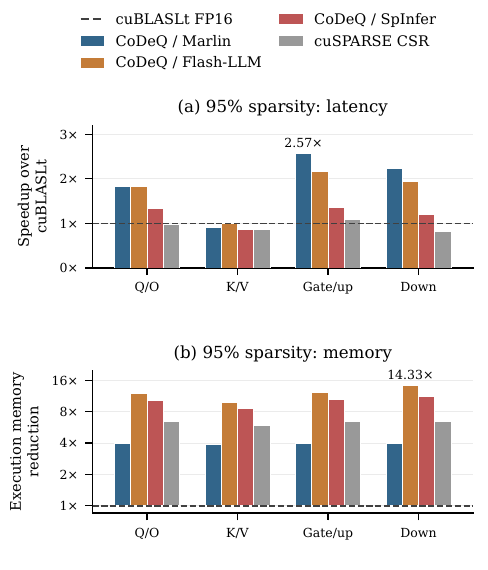}
    \vspace{-0.65cm}
    \caption{Hardware execution at 95\% target weight sparsity,
    with $T=8$ on an A100-SXM4-40GB. Here $T$ denotes the number
    of input rows processed per matrix multiplication. Across all four synthetic projection shapes, we show ratios of median latency and total execution memory relative to cuBLASLt FP16; higher is better, and the dashed line marks this baseline. 
    % All methods use identical FP16-rounded CoDeQ weights. For a full sparsity sweep and memory breakdown see Appendix~\ref{app:kernel_benchmark}.
    }  
    \label{fig:kernel_t8_focus}
    \vspace{-0.5cm}
\end{figure}

We evaluate matrix-multiplication kernels using CoDeQ fused into Marlin, Flash-LLM, and SpInfer~\cite{frantar2024marlin,xia2024flashllm,fan2025spinfer}, with cuBLASLt FP16 and cuSPARSE CSR as references. The benchmark uses an A100-SXM4-40GB, four synthetic projection shapes, six target sparsities, and $T=8$ input rows. The adapted kernels store weights in four-bit containers and compute in FP16. We measure median kernel latency and total execution memory, including weight storage, metadata, workspace, and input/output buffers. 
\Cref{fig:kernel_t8_focus} highlights 95\% target sparsity. Marlin reduces total execution memory by $3.92$--$3.97\times$ and gives $0.90$--$2.57\times$ speedups relative to cuBLASLt FP16. Flash-LLM reaches $2.16\times$ speedup and reduces memory by $9.80$--$14.33\times$; SpInfer reduces memory by $8.59$--$11.25\times$. Latency gains depend on matrix shape, with the narrow K/V projection giving little benefit. Appendix~\ref{app:kernel_benchmark} reports the full sparsity sweep, timing protocol, and memory breakdown.

%% file: files/discussion.tex
\section{Discussion}
\label{sec:discussion}
\textbf{BOPs as an idealized arithmetic measure.}
Joint pruning--quantization changes both the number of effective MACs and their operand precision. BOPs place these two axes on a common arithmetic-complexity scale and are an established evaluation metric in joint compression~\cite{joint:wang2020djpq,joint:van2020bayesian,joint:qu2025automatic,joint:FITCompress}. We follow this convention to compare sparse and quantized models independently of a particular device or kernel.

The count in \Cref{eq:bops_definition} assumes ideal zero-weight skipping at the nominal operand bit widths. Sparse data handling and decoding quantized weight add overhead outside this count. For A32, the factor 32 records activation precision without pricing floating-point arithmetic relative to integer instructions. BOPs therefore express arithmetic complexity; matching counts do not imply matching latency or energy use, and storage can also differ.\\
\textbf{Unstructured sparsity and hardware execution.}
Unstructured pruning offers flexibility in selecting individual weights, while structured pruning removes groups that are easier to exploit with conventional dense operators. In memory-bandwidth-bound regimes such as small-batch autoregressive decoding, compact sparse formats can substantially reduce weight storage and memory traffic by storing only retained values and sparse metadata~\cite{xia2024flashllm}. Our CUDA adaptations fuse four-bit codebook decoding with dense or sparse matrix multiplication~\cite{frantar2024marlin,xia2024flashllm,fan2025spinfer}, including reconstruction of the nonzero offset. They use four-bit storage containers and FP16 computation.\\
\textbf{Future work.}
Further work includes device-aware compression objectives, broader end-to-end inference evaluation, and structured or semi-structured variants that retain pruning-aware grid allocation.

%\paragraph{Compression control and weight decay.}
%The joint penalty acts on retained-range fractions and bit widths, scaled by $\lambda$. Model-weight decay instead acts on latent weights. Because $R$ tracks weight magnitudes, uniformly rescaling the weights also rescales $R$ and the threshold at fixed $\theta_{\mathrm{dz}}$, leaving the relative pruning boundary unchanged. Learning $\theta_{\mathrm{dz}}$ directly changes this boundary.

%\paragraph{Fixed-bit training and learned allocation.}
%Fixed-bit CoDeQ addresses deployments with a prescribed weight format, such as W4, chosen to match available kernels and storage layouts. Learning sparsity can then further reduce weight storage and BOPs within that format. MP adds flexibility when layerwise bit widths can be deployed.

%% file: files/conclusion.tex
\section{Conclusion}
\label{sec:conclusion}
We introduced CoDeQ, a compact QAT method that learns unstructured
weight sparsity through a parameterized dead-zone quantizer.
By reallocating nonzero reconstruction levels over the retained
range, CoDeQ supports fixed-bit and mixed-precision training under
one compression coefficient. Experiments across convolutional
networks, vision transformers, and BERT demonstrate substantial
BOP reductions while preserving task performance. With 2-bit core
activations, VGG7 achieves 0.05\% relative BOPs with accuracy only
0.32 percentage points below FP32. W4 BERT reduces BOPs by
91.55\% with a 0.63-point F1 decrease and essentially unchanged
exact match. Our kernel adaptations fuse CoDeQ codebook decoding
with dense or sparse FP16 matrix multiplication showing the potential for hardware acceleration.

% \subsection*{Acknowledgments} Authors thank members of Machine Learning Section and \hyperlink{https://saintslab.github.io/}{SAINTS Lab} for useful discussions throughout.
% Authors acknowledge the funding received from the European Union’s Horizon Europe Research and Innovation Action programme under grant agreements No. 101070284, No. 101070408 and No. 101189771. RS also acknowledges funding received under Independent Research Fund Denmark (DFF) under grant agreement number 4307-00143B.
% \clearpage

%% file: files/appendix.tex
\section*{Appendix}
\appendix

\input{files/experimental_setup}

\input{files/variational_proof}

\input{files/minimax_proof}
\input{files/backward_derivatives}
\input{files/kernel_benchmark}

%% file: files/experimental_setup.tex
\section{Experimental setup}
\label{app:experimental_setup}

\subsection{Training settings}
\Cref{tab:training_setup} summarizes the main training settings. W4 and MP share these settings. SGD uses momentum 0.9; AdamW uses $(\beta_1,\beta_2)=(0.9,0.999)$ and $\epsilon=10^{-8}$. CIFAR-10 uses epochwise cosine decay to zero without warmup. ImageNet schedules use five warmup epochs followed by cosine decay to 1\% of each initial rate, updated per optimizer step. BERT uses linear warmup over 10\% of optimizer steps. BERT accumulates four minibatches of 16 tokenized question--context features per update; the vision recipes use no accumulation. ImageNet and BERT training uses BF16 autocasting.
\textbf{CIFAR-10.}
We train ResNet-20, VGG7, and SimpleViT from scratch. All three use channel means $(0.4914,0.4822,0.4465)$ and standard deviations $(0.2470,0.2435,0.2616)$.\\
\textbf{ImageNet-1K.}
We fine-tune the non-distilled DeiT-Tiny and ViT-Small. DeiT uses channel means $(0.485,0.456,0.406)$ and standard deviations $(0.229,0.224,0.225)$; ViT-Small uses 0.5 for every mean and standard deviation.\\
\textbf{SQuAD 1.1.}
We fine-tune BERT-base-uncased. Answer decoding considers the top 20 start and end positions per window, limits answers to 30 tokens, and selects the valid span with the largest summed start/end logit across windows.

\begin{table}[!htbp]
\centering
\small
\setlength{\tabcolsep}{4pt}
\renewcommand{\arraystretch}{1.12}
\caption{Main training settings. Batch denotes the effective batch per optimizer update; LR is the initial learning rate (the peak rate after warmup). WD is model-weight decay.}
\label{tab:training_setup}
\begin{tabular*}{\textwidth}{@{\extracolsep{\fill}}lrrlrrl@{}}
\toprule
Model & Epochs & Batch & Optimizer & LR & WD & Schedule / warmup\\
\midrule
ResNet-20 & 350 & 128 & SGD & $10^{-1}$ & $10^{-4}$ & Cosine / none\\
VGG7 & 300 & 128 & AdamW & $10^{-3}$ & 0 & Cosine / none\\
SimpleViT & 100 & 128 & AdamW & $10^{-3}$ & 0.05 & Cosine / none\\
DeiT-Tiny & 100 & 512 & AdamW & $5\times10^{-5}$ & 0.05 & Cosine / 5 epochs\\
ViT-Small & 100 & 512 & AdamW & $5\times10^{-5}$ & 0.05 & Cosine / 5 epochs\\
BERT-base & 5 & 64 & AdamW & $3\times10^{-5}$ & 0.01 & Linear / 10\% of steps\\
\bottomrule
\end{tabular*}
\end{table}

\subsection{Quantization settings}
All convolutional and linear weight tensors are quantized, including the input projection and classifier or question-answering head. Biases, normalization parameters, and embedding lookup tables remain unquantized. Each weight tensor has one learned retained-range parameter and, in MP, one learned precision parameter. W4 fixes every quantized weight tensor at 4 bits; MP learns integer precisions from 2 to 8 bits using adjacent-grid feedback in \Cref{eq:adjacent}.
VGG7 uses per-tensor LSQ activation quantizers~\cite{esser2020learned}: the image input is 8-bit and other convolutional/linear inputs are 2-bit. Scales and signedness are initialized from one training minibatch. With $Q_+$ the largest positive activation code, the initial scale is $2\operatorname{mean}(|x|)/\sqrt{Q_+}$; its gradient is scaled by $1/\sqrt{NQ_+}$, where $N$ counts activation elements per example. Scales learn at the model learning rate with zero decay. 

\begin{table}[!htbp]
\centering
\small
\setlength{\tabcolsep}{4pt}
\renewcommand{\arraystretch}{1.12}
\caption{Main compression settings, shared by W4 and MP. Input and other activation bits refer to inputs of the first and remaining quantized layers, respectively, where 32 denotes no activation quantization. The precision learning rate applies only to MP.}
\label{tab:quantization_setup}
\begin{tabular*}{\textwidth}{@{\extracolsep{\fill}}lrrrrr@{}}
\toprule
Model & Input bits & Other act. bits & $\lambda$ & Retained-Range LR & Precision LR (MP)\\
\midrule
ResNet-20 & 32 & 32 & 10 & $5\times10^{-5}$ & $5\times10^{-5}$\\
VGG7 & 8 & 2 & 40 & $5\times10^{-5}$ & $5\times10^{-5}$\\
SimpleViT & 32 & 32 & 2 & $5\times10^{-5}$ & $5\times10^{-5}$\\
DeiT-Tiny & 32 & 32 & 5 & $5\times10^{-5}$ & $5\times10^{-5}$\\
ViT-Small & 32 & 32 & 5 & $5\times10^{-5}$ & $5\times10^{-5}$\\
BERT-base & 32 & 32 & 10 & $10^{-4}$ & $10^{-4}$\\
\bottomrule
\end{tabular*}
\end{table}

\subsection{Ablation settings}
\Cref{tab:ablation_setup} lists departures from the main-comparison recipes. The ResNet-20 ablations retain batch size 128, SGD momentum 0.9, model LR 0.1, compression LR $5\times10^{-5}$ and cosine decay without warmup

\begin{table}[!htbp]
\centering
\small
\setlength{\tabcolsep}{3pt}
\renewcommand{\arraystretch}{1.15}
\caption{Ablation protocols. Settings not listed here follow the corresponding main-comparison recipe.}
\label{tab:ablation_setup}
\begin{tabular}{@{}p{0.23\textwidth}p{0.18\textwidth}p{0.55\textwidth}@{}}
\toprule
Study & Base recipe & Changes or measurements\\
\midrule
Nonzero-level allocation & ResNet-20 W4 & 100 epochs, zero model WD. P+U: $\lambda\in\{0.1,5,10\}$; CoDeQ: $\lambda\in\{6.25,10,15\}$.\\
Compression pressure & ResNet-20 W4 & 100 epochs, zero model WD; vary only $\lambda\in\{1,5,6.25,10,15,18\}$.\\
Precision gradients & ResNet-20 MP & Analytic bit STE versus adjacent-grid feedback; compare the final 50 epochs of 350-epoch runs.\\
Joint trajectory & SimpleViT MP & Epochwise precision and sparsity from one run of the selected recipe.\\
Layerwise allocation & SimpleViT MP & Final-epoch layerwise measurements averaged over three runs.\\
\bottomrule
\end{tabular}
\end{table}

\paragraph{Precision gradients and compression pressure.}
\label{app:ablation_adjacent_gradients}
The precision-gradient comparison in \Cref{fig:adjacent_grid_stability} changes only the bit backward rule; both methods use $\lambda=10$. Bit-width changes are counted over the displayed epoch window.
\label{app:ablation_lambda}
The pressure sweep in \Cref{fig:compression_behavior}\subref{fig:lambda_sensitivity_w4} reports final-epoch accuracy and measured relative BOPs as mean $\pm$ sample SD over three runs at each coefficient.

\paragraph{P+U backward pass.}
\label{app:ablation_nonzero_levels}
For \Cref{tab:prune_uniform_w4}, P+U learns a magnitude threshold while leaving its nonzero uniform grid unchanged when the threshold moves at fixed $R$. Its hard-forward mask uses the sigmoid backward relaxation of Learned Threshold Pruning~\cite{azarian2020learned}:
\begin{equation}
\begin{aligned}
 m_i^H&=\mathbf1_{\{|w_i|>d/2\}},\qquad
 m_i^S=\sigma\!\left(\frac{|w_i|-d/2}{T}\right),\\
 \hat w_i&=\left[m_i^H+m_i^S-\operatorname{stopgrad}(m_i^S)\right]q(w_i;s_{\mathrm U},b),
\end{aligned}
\label{eq:pu_sigmoid_mask}
\end{equation}
where $\sigma$ is the logistic sigmoid, $s_{\mathrm U}=R/Q_b$, and $T=R/20$. Rounding and clipping use identity STEs. Both methods use the joint compression objective in \Cref{eq:objective}. Reconstruction MSE pools squared differences from clipped latent weights over surviving quantized weights within each run. Reachable levels count signed nonzero reconstruction values, averaged equally across layers, with W4 having capacity 14.

%% file: files/variational_proof.tex
\section{Reconstruction and nonzero cost: proof}
\label{app:variational_proof}
We prove Proposition~\ref{prop:assignment} for the fixed codebook of the quantizer in \Cref{eq:operator}. Its reconstruction levels are
\begin{equation}
\begin{aligned}
 \mathcal C_{s,d,b}&=\{0,\pm r_1,\ldots,\pm r_{Q_b}\},\\
 r_j&=d/2+(j-\tfrac12)s,\qquad j=1,\ldots,Q_b.
\end{aligned}
\label{eq:codebook}
\end{equation}
Set $\delta=d/2-s/2$ and
\begin{equation}
F_w(z)=\tfrac12(w-z)^2+\rho\mathbf1_{\{z\ne0\}},
\qquad \rho=(d^2-s^2)/8.
\end{equation}
\begin{proof}
First take $w\ge0$. Every positive reconstruction level $r_j$ is at least as good as its negative counterpart, since $F_w(-r_j)-F_w(r_j)=2wr_j\ge0$.
For the first positive level $r_1=d/2+s/2$, direct substitution gives
\begin{equation}
F_w(r_1)-F_w(0)=r_1(d/2-w).
\label{eq:variational_zero_boundary}
\end{equation}
If $0\le w\le d/2$, zero therefore beats or ties $r_1$. The remaining positive levels satisfy $r_j\ge r_1>d/2\ge w$, so they cannot improve its squared error and have the same nonzero cost. Thus zero is a minimizer.

If $w>d/2$, \Cref{eq:variational_zero_boundary} shows that $r_1$ already beats zero. The common cost $\rho$ cancels when comparing nonzero levels, leaving nearest-neighbor reconstruction. Adjacent positive levels have decision boundaries
\begin{equation}
\frac{r_j+r_{j+1}}2=d/2+js,
\quad j=1,\ldots,Q_b-1.
\end{equation}
For these inputs the encoder of \Cref{eq:operator} rounds $(w-\delta)/s$, whose boundaries are exactly $d/2+js$. Clipping selects the largest reconstruction level in the final unbounded quantization bin. The zero-to-first-level boundary is $w=d/2$, where zero and $r_1$ tie. For $d/2=w=0$, the convention $\operatorname{sign}(0)=0$ also returns a minimizer. At other rounding ties either adjacent output is optimal. Symmetry gives the result for $w<0$.

The vector objective is separable, so minimizing coordinatewise over the Cartesian product $\mathcal C_{s,d,b}^{\,n}$ proves \Cref{eq:assignment}.
\end{proof}
The argument applies to any actual positive step size, including a numerical stabilizer. A step size obtained from a weight-range statistic is held fixed for this characterization. The result establishes the reconstruction decision conditional on a codebook, without optimizing that codebook or establishing descent or convergence of STE training.

%% file: files/minimax_proof.tex
\section{Optimal spacing of the nonzero levels: proof}
\label{app:minimax_proof}
We prove Proposition~\ref{prop:minimax} for retained inputs with the largest reconstruction level fixed at $R$.

\begin{proof}
Let a feasible set of positive reconstruction levels have worst-case reconstruction error $E$ on $(d/2,R]$. The interval of radius $E$ around each reconstruction level covers at most length $2E$ of $(d/2,R]$. The interval around the required endpoint reconstruction level $R$ covers at most $E$ of that interval. Since these intervals must cover all retained inputs,
\begin{equation}
 R-d/2\le 2E(Q_b-1)+E=(2Q_b-1)E.
\end{equation}
For the matching construction, set $E_*=(R-d/2)/(2Q_b-1)$ and $r_j=d/2+(2j-1)E_*$, $j=1,\ldots,Q_b$. The first level is $E_*$ above $d/2$, adjacent levels are separated by $2E_*$, and the last level equals $R$. Nearest-level reconstruction therefore has supremum error $E_*$ on $(d/2,R]$. This attains the lower bound and proves \Cref{prop:minimax}. Setting $s=2E_*$ gives \Cref{eq:scale} and the positive reconstruction levels of the CoDeQ codebook.
\end{proof}
The endpoint constraint is essential: if all positive reconstruction levels are free, the optimal worst-case reconstruction error on $(d/2,R]$ is $(R-d/2)/(2Q_b)$. For signed weights the same argument applies to the negative branch by symmetry. The proof concerns reconstruction of retained inputs; pruned weights and inputs outside the chosen range contribute separate errors.

%% file: files/backward_derivatives.tex
\section{Backward derivatives of the dead-zone quantizer}
\label{app:backward_derivatives}
We derive the scalar surrogate derivatives used in \Cref{eq:combined_gradients} and compare adjacent-grid feedback with an analytic bit STE.

\subsection{Computation graph and STE rules}
Fix the detached range $R>0$ and selected bit width $b\ge2$, using the stabilized step size in \Cref{eq:training_scale}. Expanding \Cref{eq:operator} gives
\begin{equation}
\begin{aligned}
 \sigma&=\operatorname{sign}(w), & x&=|w|-\delta, & t&=\operatorname{ReLU}(x),\\
 u&=\sigma t/s, & j&=\lfloor u\rceil, & \bar w&=\operatorname{clip}(j,-Q_b,Q_b),\\
 h&=\operatorname{sign}(\bar w), & \hat w&=s\bar w+\delta h.
\end{aligned}
\label{eq:backward_graph}
\end{equation}
Write $J_w$ and $J_\xi$ for the output's backward derivatives with respect to $w$ and a quantizer parameter $\xi$, with $s_\xi=\partial s/\partial\xi$ and $\delta_\xi=\partial\delta/\partial\xi$. Absolute value uses derivative $\sigma$, with $\sigma=0$ at $w=0$; both sign operations have zero backward derivatives. Let $m_{\mathrm{ReLU}}=\mathbf1_{\{x>0\}}$ and $m_{\mathrm{clip}}=\partial_j\operatorname{clip}(j,-Q_b,Q_b)$. The clipping mask is one inside $(-Q_b,Q_b)$, zero outside, and uses the autodiff subgradient at the endpoints.

Inside a fixed quantization bin, the hard encoder is locally constant; the remaining decoder contribution gives the first row of \Cref{tab:backward_stages}. With an identity STE for rounding, the encoder chain rule is
\begin{equation}
 \frac{\partial u}{\partial w}=\frac{\sigma^2m_{\mathrm{ReLU}}}{s},
 \qquad
 \frac{\partial u}{\partial\xi}
 =-\frac{\sigma m_{\mathrm{ReLU}}\delta_\xi}{s}-\frac{u s_\xi}{s}.
\end{equation}
Differentiating the decoder and applying the standard clipping mask gives the rounding-STE row below. The ReLU and clipping STEs replace their respective masks by $1$; the clipping STE also returns zero gradients for its bounds.

\begin{table}[ht]
\centering
\small
\caption{Backward derivatives at fixed $b,R$. Every row uses the same hard forward operator; $J_\xi$ includes encoder and decoder paths.}
\label{tab:backward_stages}
\begin{tabular}{@{}lll@{}}
\toprule
Backward rules & $J_w$ & $J_\xi$\\
\midrule
No STE & $0$ & $s_\xi\bar w+h\delta_\xi$\\
Rounding STE & $\sigma^2m_{\mathrm{ReLU}}m_{\mathrm{clip}}$ & $s_\xi(\bar w-m_{\mathrm{clip}}u)+(h-m_{\mathrm{clip}}\sigma m_{\mathrm{ReLU}})\delta_\xi$\\
Rounding + ReLU STEs & $\sigma^2m_{\mathrm{clip}}$ & $s_\xi(\bar w-m_{\mathrm{clip}}u)+(h-m_{\mathrm{clip}}\sigma)\delta_\xi$\\
All three STEs & $\sigma^2$ & $s_\xi(\bar w-u)+(h-\sigma)\delta_\xi$\\
\bottomrule
\end{tabular}
\end{table}

The ReLU STE restores the encoder path where ReLU is inactive, and the clipping STE restores it for saturated weights. The final surrogate derivative with respect to the latent weight is $J_w=\sigma^2=\mathbf1_{\{w\ne0\}}$, including pruned and clipped quantized values. Exactly zero latent weights retain a zero derivative under the standard derivative of absolute value.

\subsection{Gradients for compression parameters}
Let $K_b=Q_b-\tfrac12$. For \Cref{eq:range_coordinate}, $d_{\theta_{\mathrm{dz}}}=-2R$, so $s_{\theta_{\mathrm{dz}}}=R/K_b$ and $\delta_{\theta_{\mathrm{dz}}}=-RQ_b/K_b$. Substituting into the final row of \Cref{tab:backward_stages} gives
\begin{equation}
 J_{\mathrm{dz}}
 =\frac{R}{K_b}\bigl[\bar w-u-Q_b(h-\sigma)\bigr].
\label{eq:backward_retained_range}
\end{equation}
This includes the threshold, step size, encoder, and decoder paths. At parameter bounds, the same surrogate derivative is computed before projection.

Let $\beta=b_{\max}\theta_{\mathrm{bit}}$, and use $b^-$ and $\Delta_i$ from \Cref{eq:adjacent}. For $\beta\in[k,k+1)$ with integer $b_{\min}\le k<b_{\max}$, we compare the grids at bit widths $k$ and $k+1$. At $\beta=b_{\max}$, we use $b_{\max}-1$ and $b_{\max}$. The adjacent-grid rule assigns $\widetilde\partial_\beta\hat w_i=\Delta_i$. With $g_i=\partial\mathcal L_{\rm task}/\partial\hat w_i$, the chain rule gives
\begin{equation}
 \widetilde\partial_\beta\mathcal L_{\rm task}
 =\sum_i g_i\Delta_i,\qquad
 \widetilde\partial_{\theta_{\mathrm{bit}}}\mathcal L_{\rm task}
 =b_{\max}\sum_i g_i\Delta_i.
\label{eq:backward_adjacent_bits}
\end{equation}
The selected quantization grid preserves its surrogate derivatives with respect to the latent weights and retained-range fraction and is detached from $\beta$ on this task path. The alternate quantization grid shares $R,d$ and is evaluated without an autograd graph; no derivatives pass through $\Delta_i$. The dot product represents a first-order task response in quantized-weight space. The compression penalty's bit-rounding path uses its own identity STE.

Combining these task gradients with \Cref{eq:compression_coordinate_gradients} gives \Cref{eq:combined_gradients}; fixed-bit training omits the bit update. Projection follows \Cref{eq:projection} outside autograd, with optimizer state retained at the bounds and zero weight decay on both compression parameters.

\subsection{Precision-coordinate update scaling}
\label{app:precision_update_scaling}
The coordinate change also determines the effective update rate for the unrounded bit width. For plain SGD, the chain rule gives $\widetilde\partial_{\theta_{\mathrm{bit}}}\mathcal L=b_{\max}\widetilde\partial_{\beta}\mathcal L$. Before projection, the updates therefore satisfy
\begin{equation}
\begin{aligned}
 \theta_{\mathrm{bit}}'&=\theta_{\mathrm{bit}}-\eta\,\widetilde\partial_{\theta_{\mathrm{bit}}}\mathcal L,\\
 \beta'&=\beta-\eta b_{\max}^{2}\,
                  \widetilde\partial_{\beta}\mathcal L.
\end{aligned}
\label{eq:sgd_precision_scaling}
\end{equation}
Thus learning $\theta_{\mathrm{bit}}$ at rate $\eta$ is equivalent to learning the unrounded bit width $\beta$ at rate $\eta b_{\max}^{2}$. The normalization makes this relative update scaling explicit while allowing both compression coordinates to use one learning rate.

\subsection{Comparison with the analytic bit STE}
For comparison, an analytic bit STE differentiates the step size through the bit width, using an identity STE for bit rounding and zero gradients for clipping bounds. At the selected forward $b$, using $\delta_b=-s_b/2$ in the final table row gives
\begin{equation}
\begin{aligned}
 (Q_b)_b&=(\ln2)(Q_b+1),\qquad
 s_b=-\frac{(R-d/2)(Q_b)_b}{K_b^2},\\
 J_b^{\rm STE}&=s_b\bigl[\bar w-u-\tfrac12(h-\sigma)\bigr].
\end{aligned}
\label{eq:backward_bits}
\end{equation}
The corresponding surrogate derivative with respect to $\theta_{\mathrm{bit}}$ is $b_{\max}J_b^{\rm STE}$. For the ideal step size $s=(R-d/2)/K_b>0$ in \Cref{eq:scale}, with $\epsilon=0$, let $\kappa_b=(\ln2)(Q_b+1)/K_b$. Then
\begin{equation}
 J_b^{\rm STE}=
 \begin{cases}
 -\kappa_b\operatorname{sign}(w)
       \min\{d/2-|w|,s/2\},&|w|\le d/2,\\
 \kappa_b(w-\hat w),&|w|>d/2.
 \end{cases}
\label{eq:backward_regional_bits}
\end{equation}
For zero codes, substitute $\bar w=h=0$ and $u=\sigma\max\{(|w|-d/2+s/2)/s,0\}$ into \Cref{eq:backward_bits}; for nonzero codes, use $h=\sigma$ and $u-\bar w=(w-\hat w)/s$. These substitutions establish both branches, including the zero derivative at $w=0$.

At fixed $\w,R,d$, ideal integer-bit quantizers agree on zero for $|w|\le d/2$ and on $\operatorname{sign}(w)R$ for $|w|>R$. Their adjacent-grid difference vanishes in both regions, while the analytic STE can remain nonzero. With $\epsilon>0$, the zero bin remains unchanged, but the endpoint $R+K_b\epsilon$ depends on bit width, so tail differences need not vanish.

%% file: files/kernel_benchmark.tex
\section{Kernel benchmark protocol}
\label{app:kernel_benchmark}
We measure $Y=X\hat W^\top$ with $T=8$ input rows, $X\in\mathbb R^{T\times K}$, and weight shapes $(N,K)\in\{(4096,4096),(1024,4096),(14336,4096),(4096,14336)\}$. These are synthetic projection matrices, rather than trained model checkpoints. We sample latent weights from $\mathcal N(0,2/K)$ and set the pruning threshold by the target magnitude quantile for sparsities $\{0,50,80,90,95,99\}\%$. We quantize these matrices with the W4 operator in \Cref{eq:operator} and round the reconstructions to FP16, preserving their zero support. Input entries are sampled from $\mathcal N(0,1)$ and stored in FP16. All five methods use identical reconstructed weights and FP16 inputs and outputs: dense cuBLASLt, FP16 CSR through cuSPARSE, and the CoDeQ adaptations of Marlin, Flash-LLM, and SpInfer. The packed paths decode four-bit containers within matrix multiplication.

Measurements use one NVIDIA A100-SXM4-40GB, the CUDA 12.8 compiler, CUDA runtime 12.6, and PyTorch 2.14.1+cu126. The loaded library versions are cuBLASLt 120604 and cuSPARSE 12504. Each candidate is checked against the dense reference with two independent input draws; the references are also checked at up to 128 output coordinates using CPU FP64 accumulation. We confirm the three fastest valid candidates before freezing a configuration. Dense cuBLASLt uses one selected configuration per matrix shape across sparsities. Before the full sweep, 20 T=8 correctness tests cover empty, dense, boundary, and alternating-tile patterns with both eager and CUDA-graph execution.

Candidate outputs must be finite, have relative Frobenius error at most $5\times10^{-3}$, and satisfy $|Y_{ij}-Y^{\mathrm{ref}}_{ij}|\le0.01+0.02|Y^{\mathrm{ref}}_{ij}|$ at every entry. Dense-reference checks use FP64 dot products rounded to FP16, with absolute tolerance 0.003 and relative tolerance 0.005. Both checks use the same FP16 inputs and reconstructed weights as the timed operation.

\begin{figure}[!t]
\centering
\includegraphics[width=0.90\textwidth]{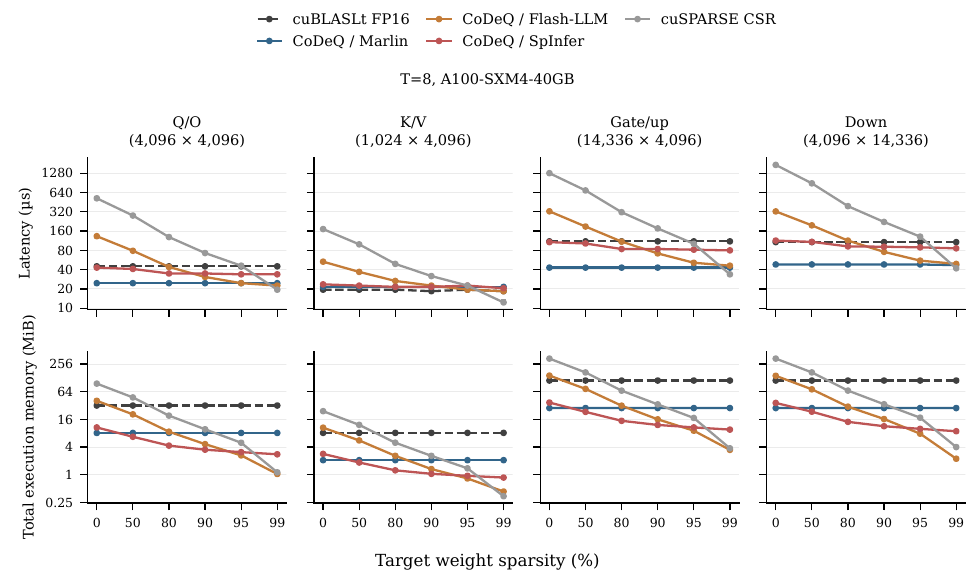}
\caption{Kernel latency and total execution memory at $T=8$ on an A100-SXM4-40GB. All methods use the same FP16-rounded W4 CoDeQ reconstructions for each synthetic matrix. Top: median latency, with sample 10th--90th percentiles over 300 measurements. Bottom: stored-weight memory (including codebooks, sparse metadata, and padding), workspace, and input/output. Both vertical axes are logarithmic; horizontal positions denote the six tested sparsity settings.}
\label{fig:kernel_t8}
\end{figure}

Final timings comprise three rounds of 100 samples per method/case, with shuffled method order, 20 warmups, and 100 ms settling. Before each sample, we write a 160 MiB buffer to evict L2 contents, outside the timed interval. CUDA-graph event timings cover matrix multiplication and required GPU conversions and reductions; direct CUDA-event timings provide a cross-check. Packing, uploads, allocation, and tuning occur before timing. Figure~\ref{fig:kernel_t8} reports median latency and sample 10th--90th percentiles. Execution memory comprises stored weights, codebooks, sparse metadata, padding, workspace, and input/output buffers for each backend. Benchmark-only reference and cache-eviction buffers are excluded. Table~\ref{tab:kernel_t8_memory} separates stored-weight memory from workspace.

%At 99\% target sparsity, the Flash-LLM adaptation reduces stored-weight memory by $77\times$ relative to cuBLASLt FP16 across all four shapes. Reductions in total execution memory range from $18.88$ to $50.25\times$ after including workspace and input/output. For the Down projection, stored-weight memory decreases from $112$ to $1.453$ MiB ($77.07\times$), while total execution memory decreases from $112.281$ to $2.235$ MiB ($50.25\times$). The difference reflects $0.500$ MiB of workspace and $0.281$ MiB of input/output in the compressed path.

\input{files/table_kernel_t8_memory}

%% file: files/table_kernel_t8_memory.tex
\begin{table}[!t]
\centering
\small
\setlength{\tabcolsep}{6pt}
\caption{T=8 memory accounting on the A100-SXM4-40GB. Ranges span the six tested sparsities. Stored-weight memory includes codebooks, sparse metadata, and padding. Total execution memory also includes workspace and input/output.}
\label{tab:kernel_t8_memory}
\begin{tabular}{llrrr}
\toprule
Shape & Method & \makecell{Weights + metadata\\(MiB)} & \makecell{Workspace\\(KiB)} & \makecell{Total\\(MiB)}\\
\midrule
Q/O & cuBLASLt FP16 & 32.00 & 0.00 & 32.12\\
Q/O & CoDeQ / Marlin & 8.00 & 0.12 & 8.13\\
Q/O & CoDeQ / Flash-LLM & 0.42--40.01 & 512.00 & 1.04--40.63\\
Q/O & CoDeQ / SpInfer & 2.16--10.08 & 512.00 & 2.79--10.70\\
Q/O & cuSPARSE CSR & 0.98--96.02 & 32.02 & 1.13--96.17\\
\midrule
K/V & cuBLASLt FP16 & 8.00 & 80.00 & 8.16\\
K/V & CoDeQ / Marlin & 2.00 & 0.03 & 2.08\\
K/V & CoDeQ / Flash-LLM & 0.10--10.00 & 256.00--512.00 & 0.43--10.58\\
K/V & CoDeQ / SpInfer & 0.54--2.52 & 256.00 & 0.87--2.85\\
K/V & cuSPARSE CSR & 0.24--24.00 & 20.02--100.02 & 0.34--24.18\\
\midrule
Gate/up & cuBLASLt FP16 & 112.00 & 0.00 & 112.28\\
Gate/up & CoDeQ / Marlin & 28.00 & 0.22--0.44 & 28.28\\
Gate/up & CoDeQ / Flash-LLM & 1.45--140.03 & 1792.00--3584.00 & 3.48--143.81\\
Gate/up & CoDeQ / SpInfer & 7.58--35.27 & 1792.00 & 9.61--37.30\\
Gate/up & cuSPARSE CSR & 3.41--336.05 & 72.02--360.02 & 3.77--336.69\\
\midrule
Down & cuBLASLt FP16 & 112.00 & 0.00 & 112.28\\
Down & CoDeQ / Marlin & 28.00 & 0.06--0.12 & 28.28\\
Down & CoDeQ / Flash-LLM & 1.45--140.03 & 512.00--2048.00 & 2.23--142.31\\
Down & CoDeQ / SpInfer & 7.58--35.27 & 1024.00--2048.00 & 8.86--36.55\\
Down & cuSPARSE CSR & 3.38--336.02 & 360.02 & 4.01--336.65\\
\bottomrule
\end{tabular}
\end{table}